\documentclass[pdflatex,sn-mathphys-num]{sn-jnl}%
\usepackage{graphicx}
\usepackage{multirow}
\usepackage{amsmath,amssymb,amsfonts}
\usepackage{amsthm}
\usepackage{mathrsfs}
\usepackage[title]{appendix}
\usepackage{xcolor}
\usepackage{textcomp}
\usepackage{manyfoot}
\usepackage{booktabs}
\usepackage{algorithm}
\usepackage{algorithmicx}
\usepackage{algpseudocode}
\usepackage{listings}
\usepackage{float}
\usepackage{color}
\usepackage{tikz}
\usepackage{threeparttable}
\usepackage{subcaption}
\usepackage[utf8]{inputenc}
\usepackage{xcolor}
\usetikzlibrary{fit}
\usetikzlibrary{backgrounds}
\usetikzlibrary{matrix}
\usetikzlibrary{decorations.pathmorphing}
\usetikzlibrary{decorations.markings}
\usetikzlibrary{calc}
\usetikzlibrary{arrows,shapes,positioning,shadows,trees}

\theoremstyle{thmstyleone}%
\theoremstyle{thmstyletwo}%

\theoremstyle{thmstylethree}%

\begin{document}

\title{Diffusion-Driven Deceptive Patches: Adversarial Manipulation and Forensic Detection in Facial Identity Verification}

\author*{\fnm{Shahrzad} \sur{Sayyafzadeh}}\email{shahrzad1.sayyafzade@famu.edu}

\author{\fnm{Hongmei} \sur{Chi}}\email{hongmei.chi@famu.edu}
\author{\fnm{Shonda} \sur{Bernadin}}\email{bernadin@eng.famu.fsu.edu}

\affil*[1,3]{\orgdiv{Electrical \& Computer Engineering}, \orgname{FAMU-FSU College of Engineering}, \orgaddress{\street{Pottsdamer St}, \city{Tallahassee}, \postcode{32310}, \state{Florida}, \country{USA}}}

\affil[2]{\orgdiv{Department of Computer \& Information Science},\\ \orgname{Florida A\&M University},\orgaddress{\street{1333 Wahnish Way}, \city{Tallahassee}, \postcode{32307}, \state{Florida}, \country{USA}}}


\abstract{This work presents an end-to-end pipeline for generating, refining, and evaluating adversarial patches to compromise facial biometric systems with forensic analysis and security testing applications. We utilize a FGSM to generate adversarial noise targeting our classifier for identity detection and employ a diffusion model for reverse diffusion to enhance the imperceptibility with additional Gaussian smoothing and adaptive brightness correction of synthetic adversarial patch evasion generation. The refined patch is applied to facial images to test its ability to evade recognition systems while maintaining natural visual characteristics. A Vision Transformer (ViT)-GPT2 model generates captions to provide a semantic description of a person’s identity for Adv Images, supporting forensic interpretation and documentation for identity evasion attack and recognition. The pipeline evaluates changes in identity classification, captioning results, and the vulnerability of facial identity verification and expression to adversarial attacks. Therefore, detecting and mitigating attacks from these adversaries is necessary in forensic settings using perceptual hashing. We successfully detected and analyzed a series of adversaries generated with 0.95\% SSIM. }




\maketitle

\section{Introduction}\label{sec1}
Deep learning models have revolutionized image classification \cite{hwang2023adversarial}, achieving remarkable accuracy in computer vision applications, from facial recognition to medical diagnostics. However, these models exhibit critical vulnerabilities to adversarial attacks and carefully crafted perturbations that can deceive classifiers while remaining imperceptible to humans \cite{tampubolon2024digital}. One particularly susceptible area is facial emotion classification and identity recognition, where models detect emotions such as happiness, anger, sadness, and surprise and change a person's identity through the description of its respected forensic biometric systems. Adversarial patch attacks pose a concern in facial identity verification, creating localized regions of adversarial modifications that can be physically deployed. For facial images $x \in \mathbb{R}^{H \times W \times C}$. Where $H,W$ denote dimensions and $C$ represents color channels, and with a standard input size of $224 \times 224$ pixels , an adversarial patch $p$ maximizes the classifier's loss:

\[
   \max_{p} L(f(x+p), y)
\]

where $f$ is the identity classifier and $y$ is the true emotion and gender identity label.

In addition to emotion classification, facial identity verification systems are equally vulnerable to adversarial attacks, posing severe security risks. Facial identity verification, widely used in biometric authentication and surveillance, matches facial features to stored profiles to confirm identities. Adversarial patches targeting these systems can cause false positives, allowing imposters to bypass authentication, or false negatives, rejecting legitimate users. Such manipulations exploit the deep feature embeddings on which identity verification systems rely, causing the model to misclassify identities even when the alterations are visually subtle. For identity verification models such as ArcFace and FaceNet, which commonly process facial images of size $112 \times 112$ or $160 \times 160$ pixels, these adversarial perturbations significantly degrade recognition accuracy \cite{bib1}.

To detect these manipulations, forensic analysis techniques are critical in exposing adversarial artifacts hidden from human perception. Spectral analysis using fast fourier transform (FFT) can reveal high-frequency perturbations that characterize adversarial patches, highlighting discrepancies between authentic and adversarial images. Additionally, depth estimation, achieved using MiDaS models on images resized to $256 \times 256$ pixels, can identify geometric inconsistencies introduced by adversarial attacks, such as distorted facial structures or unnatural surface textures. These forensic approaches can detect adversarial tampering that is imperceptible in the spatial domain, providing an additional layer of security against attacks on both facial emotion and identity recognition systems.

Diffusion models play a dual role in adversarial manipulation and defense. In adversarial patch generation, diffusion models employ a forward process by adding Gaussian noise over $T$ time steps, defined as:

\[
   q(x_t|x_{t-1}) = \mathcal{N}(x_t; \sqrt{1-\beta_t}x_{t-1}, \beta_t\mathbf{I})
\]

Where $\beta_t$ represents the variance schedule. The reverse process generates images from noise using:

\[
   p_\theta(x_{t-1}|x_t) = \mathcal{N}(x_{t-1}; \mu_\theta(x_t,t), \sigma_\theta^2(t)\mathbf{I})
\]

By integrating adversarial objectives into the diffusion process, the total loss is defined as:

\[
   L_{total} = L_{diffusion} + \lambda L_{adv}
\]

Where $L_{diffusion}$ ensures high-fidelity image generation, and $L_{adv}$ encourages perturbations that mislead the target classifier. This formulation allows diffusion models to generate highly effective adversarial patches that blend seamlessly into facial features, such as cheeks or foreheads, making them difficult to detect through traditional forensic techniques. The adversarial patches are typically generated on images resized to $224 \times 224$ pixels to match the input dimensions of commonly targeted models like InceptionResnetV1 pretrained on VGGface2.

Conversely, diffusion models also serve as a defense mechanism through adversarial purification. In this process, an adversarial image is passed through a reverse diffusion model to remove perturbations and restore the original image. The purified image $\hat{x}$ is reconstructed from Gaussian noise as:

\[
   \hat{x} = p_\theta(x_0|x_T) \quad \text{where} \quad x_T \sim \mathcal{N}(0,\mathbf{I})
\]

This process effectively maps adversarial images back to the natural data manifold, preserving emotion-relevant and identity-relevant features while eliminating adversarial distortions. Adversarial purification is typically performed on images resized to the model’s input dimensions, such as $112 \times 112$ for identity verification and $224 \times 224$ for emotion recognition models. By comparing purified images with their originals through forensic analysis, subtle differences in spectral and depth domains can be identified, further improving adversarial detection capabilities.

\usetikzlibrary{shapes.geometric, arrows, positioning, calc}

\tikzstyle{process} = [rectangle, rounded corners, minimum width=1.8cm, minimum height=0.6cm, text centered, draw=black, fill=blue!10, font=\tiny]
\tikzstyle{feature} = [rectangle, rounded corners, minimum width=1.8cm, minimum height=0.6cm, text centered, draw=black, fill=orange!10, font=\tiny]
\tikzstyle{decision} = [diamond, minimum width=2cm, minimum height=1cm, text centered, draw=black, fill=yellow!20, font=\tiny]
\tikzstyle{output} = [rectangle, minimum width=1.6cm, minimum height=0.5cm, text centered, draw=black, fill=green!20, font=\tiny]
\tikzstyle{caption} = [rectangle, minimum width=1.8cm, minimum height=0.6cm, text centered, draw=black, fill=purple!10, font=\tiny]
\tikzstyle{arrow} = [thick,->,>=stealth]


\subsection{Image Classification, Captioning, and Identity Verification Pipeline}
The diagram presents a pipeline for image classification, captioning, and identity verification, showcasing the effect of adversarial patches on deep learning models. An input image \(x\) is processed using ViT-GPT2, producing an initial caption \(C_{\text{original}}\). An adversarial patch \(p\) is applied to the image, generating \(x+p\), which undergoes the same classification and captioning.

Both original and patched images are processed through InceptionResnetV1 to extract class features \(f_c(x)\) and identity features \(f_i(x)\). The class features determine the predicted label, while identity features capture embeddings used for biometric verification. Meanwhile, the ViT-GPT2 model generates the patched caption \(C_{\text{patched}}\).

The pipeline compares outputs to measure the patch's impact. Class changes are detected using the L2 distance between class embeddings:
\[
\|f_c(x)-f_c(x+p)\|_2,
\]
and flagged if exceeding a threshold \(\tau_c\). Identity changes are similarly evaluated:
\[
\|f_i(x)-f_i(x+p)\|_2,
\]
with a threshold \(\tau_i\). Caption differences result from directly comparing \(C_{\text{original}}\) and \(C_{\text{patched}}\).

This pipeline reveals how adversarial patches affect classification, captioning, and identity verification, highlighting vulnerabilities in multimodal deep learning models for computer vision and biometric security in Fig.1.
\begin{figure}[h!]
    \centering
    \includegraphics[width=0.9\textwidth, height=0.4\textheight]{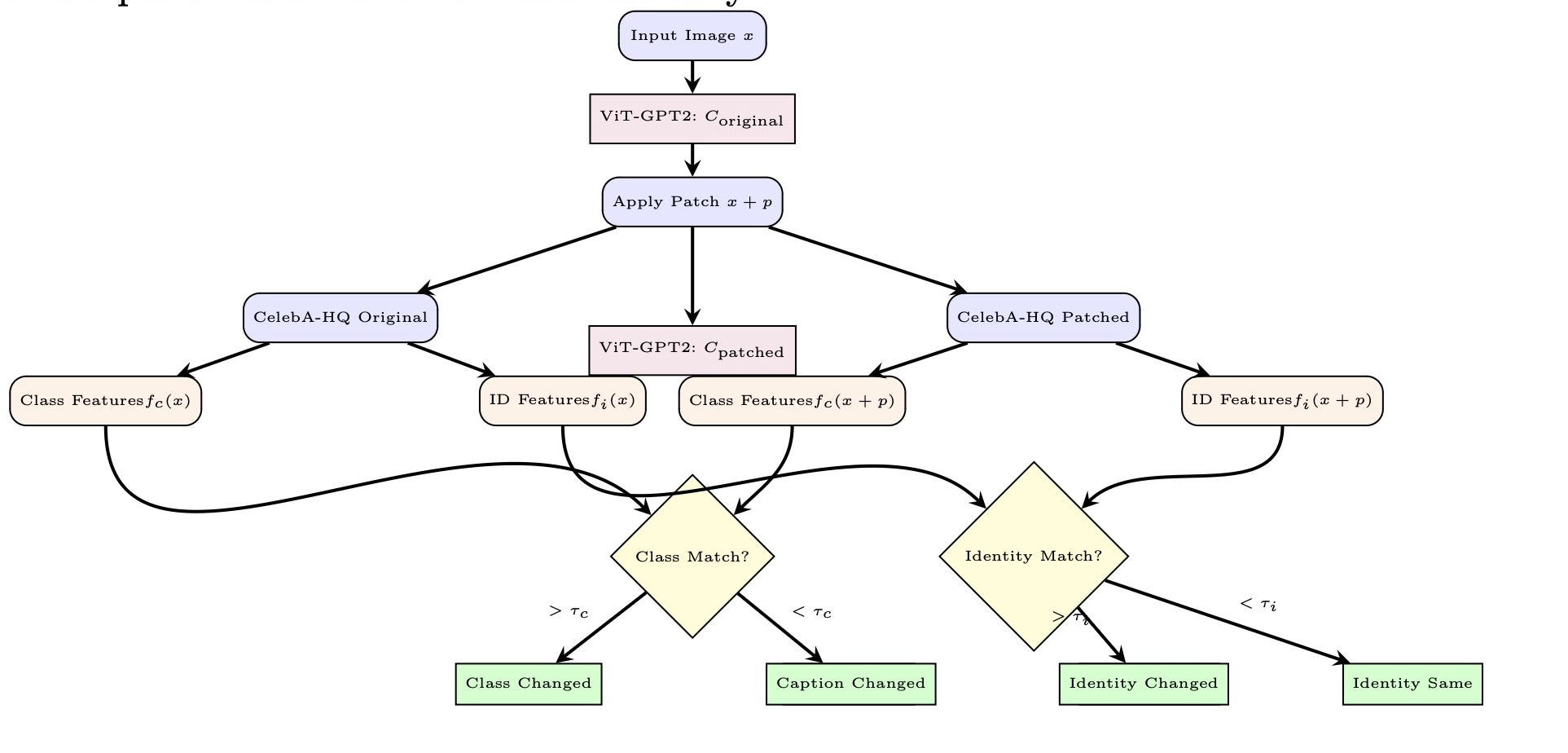}
    \caption{Illustrating a ViT-GPT2 based biometric model for status drift captioning and tracking.}
    \label{fig:surprise_expression}
\end{figure}
\subsection{Adversarial Patch Attack Detection and Digital Forensic}
The fig.1.represents the process for detecting adversarial patch attacks using perceptual hash distance in the context of digital forensics \cite{yutia2024digital}. The workflow begins with the original image processing, where the image is first converted to grayscale. After this, hashes for the image were generated using multiple hash functions, including aHash (average), pHash (perceptual), dHash (difference), and wHash (wavelet). These hashes capture the various perceptual features of the original image.
A similar process occurs for the patched image. The patched image is also converted to grayscale and hashed using the same set of hash functions. This allows the system to compare the perceptual differences between the original and patched hashes.
The system then calculates the Hamming distance between the original and patched hashes. The Hamming distance is used to measure the difference between the two hashes, with higher values indicating more perceptual differences, which may suggest the presence of an adversarial patch.
Once the Hamming distance is calculated, the system proceeds to the detection decision. A decision rule is applied: an adversarial patch is detected if any of the Hamming distances exceeds a threshold of 5. This threshold is used to determine whether the perceptual change in the image is enough to be considered tampering or manipulation.
Additionally, the system includes multimodal detection, meaning that detection is not solely reliant on hash distance. The detection process also considers other methods, such as detecting SSIM differences (Structural Similarity Index), segment anomalies, contour detection, heatmap analysis, and even label changes during classification. If any of these methods indicates the presence of an adversarial attack, the system will flag it as detected.
Therefore, this workflow outlines our proposed method for detecting adversarial patches by leveraging perceptual hashing, hamming distance calculation, and multimodal detection strategies. This approach is highly effective for identifying subtle manipulations in images, particularly when traditional image comparison methods may fail in Fig.2.
\begin{figure}[h!]
    \centering
    \includegraphics[width=0.9\textwidth, height=0.3\textheight]{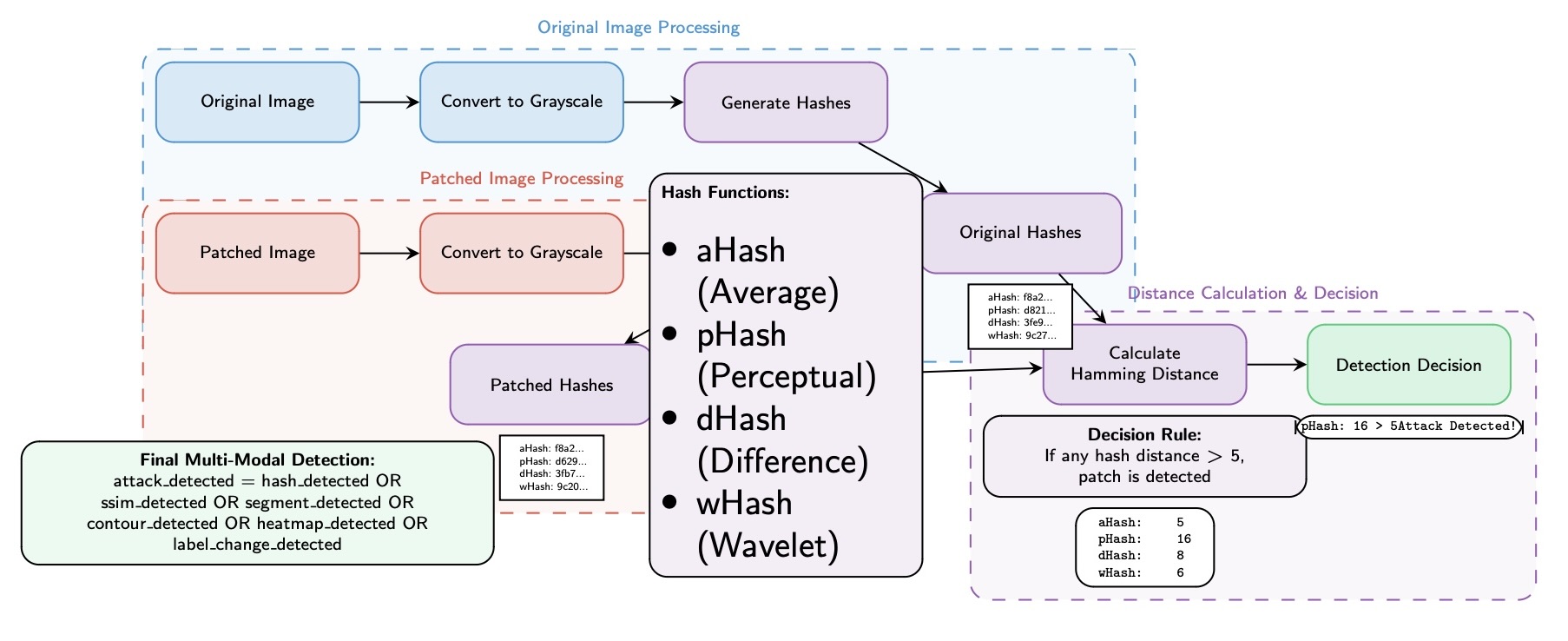}
    \caption{Demonstrating adversarial patch detection pipeline workflow and decision logic.}
    \label{fig:surprise_expression}
\end{figure}
\section{Background}\label{sec2}
Digital forensics is a critical field in cybersecurity that focuses on identifying, collecting, preserving, and analyzing digital evidence. With the increasing reliance on digital platforms, cyber threats have evolved, requiring advanced forensic techniques to detect and mitigate malicious activities. Among these threats, adversarial attacks, particularly patch attacks, have emerged as a challenge to forensic investigations and security systems \cite{shetty2020digitalforensics}. 

Adversarial patch attacks are a subset of adversarial machine learning techniques where attackers introduce carefully crafted perturbations into images, leading machine learning models to misclassify objects. These patches, generated through gradient-based optimization, stable diffusion, or adversarial attack techniques such as FGSM and PGD, can deceive deep learning models into incorrect predictions. These attacks pose severe risks in security-critical applications such as facial recognition, autonomous driving, medical imaging, and biometric authentication.

In facial recognition systems, attackers can use adversarial patches to evade identification \cite{wu2024sam,thys2019surveillance}, while in autonomous vehicles, minor alterations to traffic signs can mislead vision-based navigation, leading to potential accidents \cite{geng2023vehicle,hingun2022reap}. Similarly, in medical imaging, adversarial perturbations can cause AI models to misdiagnose diseases, and in surveillance systems, these attacks can bypass security measures, allowing unauthorized individuals to evade detection \cite{li2024pgpatch,jutras2022physical}.

Given the increasing importance of such attacks, adversarial detection mechanisms are essential to maintain the integrity of digital forensics. Hashing techniques play a crucial role in forensic analysis by detecting tampering in digital evidence. Traditionally, digital forensics relies on cryptographic and perceptual hashing to verify the integrity of digital files. Cryptographic hashing, such as SHA-256 and MD5 \cite{sayyafzadeh2023thumbnails}, generates a unique fingerprint of a file, ensuring its integrity. However, even minor pixel changes can result in drastically different hash values, making cryptographic hashing too sensitive for detecting adversarial patch attacks.

On the other hand, perceptual hashing is designed to identify visual similarities between images, making it more suitable for detecting subtle adversarial modifications. Techniques like aHash, pHash, dHash, and wHash can detect structural changes in an image without being overly sensitive to minor alterations such as compression or noise. Unlike deep learning-based adversarial detection methods that require continuous retraining and large-scale datasets, hashing-based forensics provides a lightweight and efficient alternative. Hash comparisons require minimal computational resources while still effectively identifying adversarial manipulations. This method is resilient against minor modifications, making it ideal for real-time forensic applications requiring rapid image integrity verification \cite{arvinte2020residuals,sharma2022survey}.

Law enforcement agencies, security analysts, and digital forensic investigators can integrate hash-based detection into their workflows to quickly assess whether an image has been altered in a way that could impact an investigation. These attacks manipulate images or video data to deceive machine learning models into misclassification, raising serious concerns in identity verification, biometric security, and digital authentication systems.

Identity verification and biometric verification systems are integral to modern security infrastructures. These systems leverage unique physiological or behavioral characteristics such as facial features, fingerprints, iris patterns, and voice recognition to authenticate individuals. Facial recognition, in particular, has gained widespread adoption in law enforcement, banking, border control, and mobile device authentication. However, the increasing sophistication of adversarial attacks poses a severe threat to these biometric systems \cite{mathov2021realworld}. 

Attackers can introduce imperceptible adversarial patches into images to fool recognition models into misidentifying individuals or bypassing authentication entirely. In law enforcement and surveillance applications, such manipulations could allow criminals to evade detection, while in banking and secure access systems, they could facilitate unauthorized access to sensitive information. Therefore, we investigate the opportunities for adversarial attacks to test and measure the robustness of biometric systems and leverage hash values for more accurate detection in a forensic setting.

\section{Related Work}\label{sec3}

The vulnerability of facial expression recognition (FER) systems to adversarial attacks has become a critical area of research in computer vision and security. Adversarial examples, first introduced by Goodfellow et al. \cite{Goodfellow2015}, revealed how deep learning models could be easily deceived by minimal perturbations. Building on this concept, Brown et al. \cite{Brown2017} proposed adversarial patches —local regions of perturbations capable of fooling classifiers without altering the entire image. Unlike traditional perturbations, adversarial patches are physically realizable and effective even under real-world conditions, such as camera distortions and lighting changes \cite{Sharif2019}.

In the context of facial expression recognition, Li et al. \cite{Li2021} demonstrated that adversarial patches could cause misclassification of emotions, such as interpreting anger as happiness or sadness as neutrality. Zhang et al. \cite{Zhang2022} extended these findings by exploring the real-world implications of adversarial attacks on emotion detection systems used in surveillance and driver monitoring. Their results emphasized the urgent need for defenses capable of mitigating such attacks.
  
Diffusion models have emerged as a powerful tool in both adversarial attack generation and defense. Ho et al. \cite{Ho2020} introduced denoising diffusion probabilistic model (DDPM), which demonstrated state-of-the-art performance in image synthesis by iteratively refining Gaussian noise. Song et al. \cite{Song2021} further extended diffusion models to score-based generative modeling using stochastic differential equations (SDEs), enabling controllable and flexible image generation. These advancements paved the way for using diffusion models in adversarial research, including the generation of adversarial patches and purification-based defenses \cite{Nie2022}.

Choi et al. \cite{Choi2023} proposed one of the earliest methods for generating adversarial patches using diffusion models. By integrating classifier gradients into the reverse diffusion process, they produced highly effective adversarial patches that could deceive facial recognition systems while maintaining high visual fidelity. Their approach demonstrated that diffusion-based patches were more resilient to common defenses such as JPEG compression and Gaussian noise compared to traditional gradient-based attacks. This work directly influenced research on adversarial patches for FER systems, highlighting their effectiveness in causing emotion misclassification.

Diffusion models are also effective as a defense mechanism against adversarial perturbations. Nie et al. \cite{Nie2022} proposed DiffPure, a defense framework that utilizes diffusion models to remove adversarial noise through a reverse diffusion process. Their experiments showed that DiffPure could significantly improve the robustness of classifiers against various attacks, including adversarial patches. This approach has been recognized as a promising solution for enhancing the security of FER systems, where accurate emotion detection is critical.

Despite advancements in both attacks and defenses, the intersection of diffusion models and adversarial patches for facial expression recognition remains underexplored. Existing works such as those by Li et al. \cite{Li2021} and Zhang et al. \cite{Zhang2022} have primarily focused on gradient-based attacks, leaving diffusion-based approaches largely unaddressed. Additionally, while diffusion models have shown strong capabilities in image generation and purification, their potential for generating deceptive patches specifically targeting emotion recognition models has not been fully realized.

\section{Methodology}\label{sec3}
\begin{figure}[t!] 
    \centering
    \includegraphics[width=\textwidth, height=0.9\textheight, keepaspectratio]{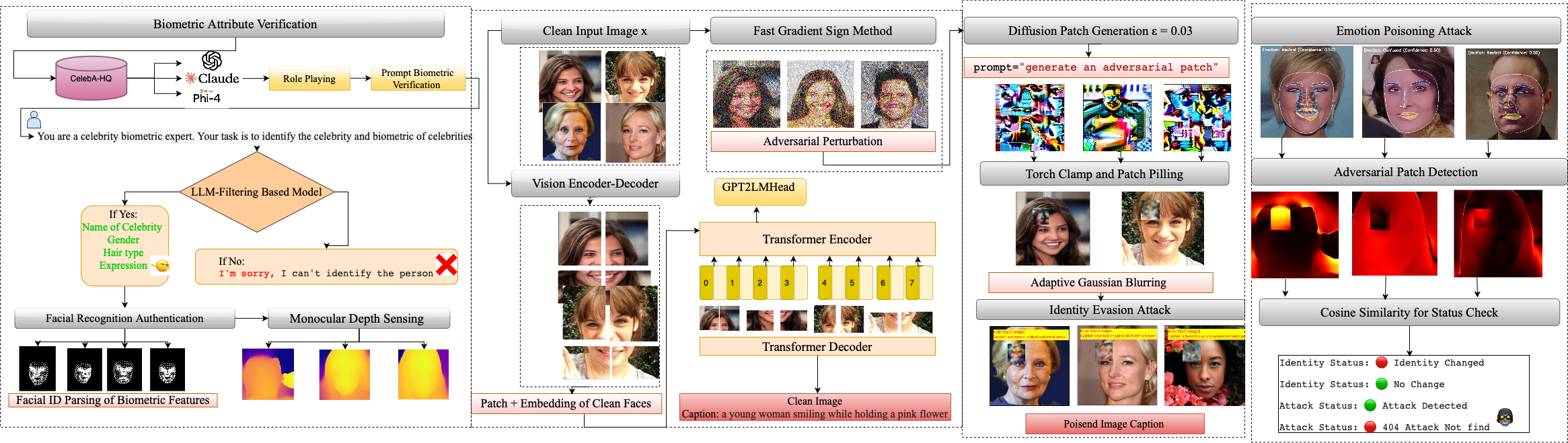}
    \caption{Using source CelebA images of 178×218 pixels and adversarial patches of 50×50 pixels, the adversarial attack pipeline demonstrates the interaction between perturbation, diffusion-based patch generation, and identity evasion. This process integrates adversarial noise with patch refinement, using cosine similarity measurements to monitor attack effectiveness report on biometric identity verification system.}
    \label{fig:diagram0}
\end{figure}

Our approach involves a multistage pipeline to analyze and mitigate adversarial attacks on AI vision models, leveraging diffusion models, adversarial patch generation, and identity evasion techniques. The methodology is structured into five key phases: adversarial perturbation, diffusion-based patch generation, identity evasion, emotion poisoning attacks, and adversarial patch detection, as depicted in Fig.3
\subsection{Biometric Attribute Verification}\label{subsec2}
In our proposed biometric attribute verification pipeline, we designed a system to identify celebrities through facial recognition and biometric analysis. The process begins with referencing a large-scale celebrity biometric dataset, such as CelebA-HQ, which contains high-resolution images of various celebrities along with their associated biometric attributes. These attributes include not only facial features but also key metadata like the name, gender, hair type, and facial expressions. This database serves as the foundation for the model, providing a rich set of data for training and comparison during the identification process.

The system interacts with the user, who assumes the role of a celebrity biometric expert. In this role, the expert is tasked with identifying the celebrity in question and verifying the biometric features associated with that individual. The system prompts the user to verify specific biometric attributes of the celebrity, such as their gender, hair type, or facial expression. This step simulates real-world scenarios of celebrity recognition and attribute verification, where the user confirms these attributes based on the extracted data from the database.

The central component of the system is the LLM-filtering based model, which uses LLM used to recognize and filter relevant celebrity features. This model processes inputs like celebrity name, gender, hair type, and expression, and then matches these attributes with the data in the dataset as a labeling method. 
The decision-making process in this model follows a logical flow: if the system successfully matches the celebrity's name with the provided attributes (gender, hair type, and expression), it proceeds with the verification of the biometric attributes. If no match is found, the system responds by saying, "I'm sorry, I can't identify the person," indicating that the celebrity could not be identified based on the celebrity identity.

Following this, facial recognition authentication is applied to validate the identified celebrity. Facial ID parsing extracts key facial features from the input image, such as the position of facial landmarks, eye shape, nose width, mouth curvature, and other unique characteristics. These features are passed through a facial recognition model, to compare the parsed facial data with the biometric database. This step is essential for ensuring that the celebrity's face is accurately identified.

Once the facial recognition is completed, the system performs biometric feature verification by cross-checking the celebrity's attributes, such as name, gender, hair type, and expression, with the verified facial features. This ensures the accuracy of both the celebrity identification and the biometric feature verification process.

A key feature of this system is its use of monocular depth sensing. This technique estimates depth information from a single image, offering more improved analysis of the face. Unlike stereo vision, which requires two images, monocular depth sensing uses computer vision algorithms, to estimate the relative distance between the camera and different objects in the scene. This method provides a 3D structure of the face from a 2D image, adding an additional layer of security to the celebrity identification process. By capturing detailed depth information, monocular depth sensing helps to distinguish between similar-looking individuals or detect any artificial manipulation of the image, such as adversarial attack.

Following monocular depth sensing, the system applies 3D face mapping, which generates a three-dimensional model of the face. This model provides a detailed understanding of the celebrity’s facial structure, including the contours of the chin, nose bridge, and forehead shape, which are crucial for accurate biometric verification. 3D face mapping adds another layer of accuracy, ensuring that the celebrity is correctly identified, even when using images with subtle variations in lighting or angles.

Last and for most of this multistep process,system consolidates the findings from facial recognition authentication, biometric feature verification, monocular depth sensing, and 3D face mapping. It then provides the final celebrity identification result, confirming the identity of the celebrity and their biometric attributes. If any step fails, the system apologizes and states that it cannot identify the person demonstrated in Fig.4.

\begin{figure}[h!]
    \centering
    \includegraphics[width=0.9\textwidth, height=0.1\textheight]{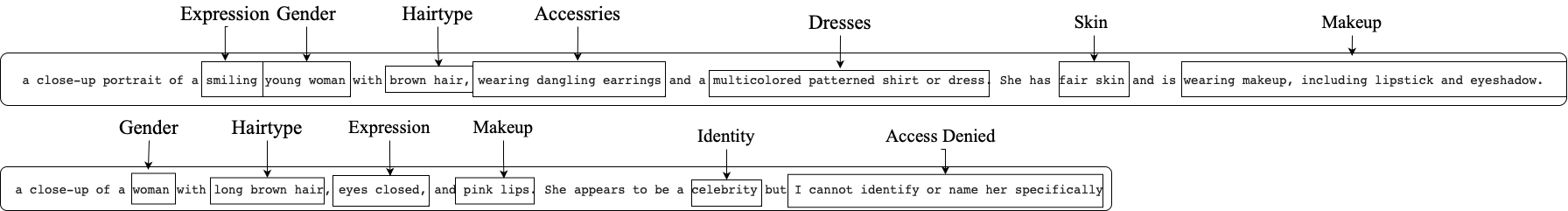}
    \caption{Analysis of biometric labeling in the context of LLM filtration using respected feedback prompts into different categories.}
    \label{fig:surprise_expression}
\end{figure}

\subsection{FSGM-Diffusion Patch Generation}\label{subsec2}
The FGSM is employed to generate adversarial perturbations by modifying an input image tensor $x$ in a direction that maximizes the model's prediction error while maintaining imperceptible visual alterations. Given an input image $x$, a target class $y$, and a perturbation magnitude $\epsilon$, the adversarial image $x'$ is calculated by adding a small perturbation $\eta$ to $x$, where $\eta$ is derived from the loss function gradient with respect to the input.

The process begins with gradient computation, where the input tensor $x$ is cloned, detached, and marked as requiring gradients to enable backpropagation. This ensures that the perturbation can be efficiently computed during the attack process. Once the input image is prepared, a forward pass is performed by feeding $x$ into a pre-trained vision model to obtain logits, which represent the model's predicted class probabilities. The adversarial loss is then calculated using the cross-entropy loss function, which quantifies the difference between the model’s output and the target class $y$. After computing the loss, backpropagation is applied to determine the gradient $\nabla_x J(\theta, x, y)$, which indicates the optimal direction in which the input image should be modified to maximize the model’s prediction error.

Once the gradient is obtained, the adversarial noise $\eta$ is generated by taking the element-wise sign of the gradient and scaling it by $\epsilon$. This ensures that the perturbation is small but effective in causing misclassification. The perturbed image is then clamped to maintain valid pixel values within the range $[0,1]$, ensuring that the modifications do not introduce unnatural artifacts. The final adversarial image $x' = x + \eta$ retains its perceptual similarity to the original image while deceiving the model into making incorrect predictions.
The refinement process of the adversarial patch incorporates a diffusion-based denoising step to improve its stealthiness while preserving its adversarial effectiveness. Starting with an initial adversarial patch \( P \), we apply a reverse diffusion process to ensure that the patch remains effective in misleading the model while appearing imperceptible to human observers. After refinement, the patch is placed onto the target image \( I \) at a specific coordinate \( (x, y) \). The refined patch \( P' \) is then resized to a smaller region \( P'' \) and further smoothed using Gaussian blurring to enhance seamless blending.  

The final adversarial image \( I' \) is obtained by integrating the smoothed patch \( P'' \) into the original image at location \( (x, y) \). This process is mathematically represented as:

\[
I' = I \odot M + P'' \odot (1 - M)
\]

Where \( M \) is a binary mask that determines the placement of the patch, and \( \odot \) denotes element-wise multiplication to ensure smooth and natural integration. The Gaussian filter \( G_{\sigma} \), with standard deviation \( \sigma \), is applied to regulate the sharpness of the patch, making it less detectable. As a result, the adversarial image \( I' \) maintains its adversarial properties while ensuring the modification remains visually indistinguishable. To evade identity recognition systems, we apply an identity evasion attack to clean images. This step ensures seamless patch integration, reducing the detectability of the attack while manipulating vision encoder-decoder embeddings. The result is a perturbed image that modifies the AI-generated caption output, leading to poisoned image captions. Beyond identity evasion, we introduce emotion poisoning attacks, where adversarial patches alter emotion recognition outputs. The attack modifies facial features, leading to incorrect emotional classifications. This phase demonstrates the broader impact of adversarial attacks beyond identification, affecting high-level semantic interpretations in AI systems in Fig.5.
\begin{figure*}[t!]
    \centering
    \includegraphics[width=0.80\textwidth, height=0.50\textheight]{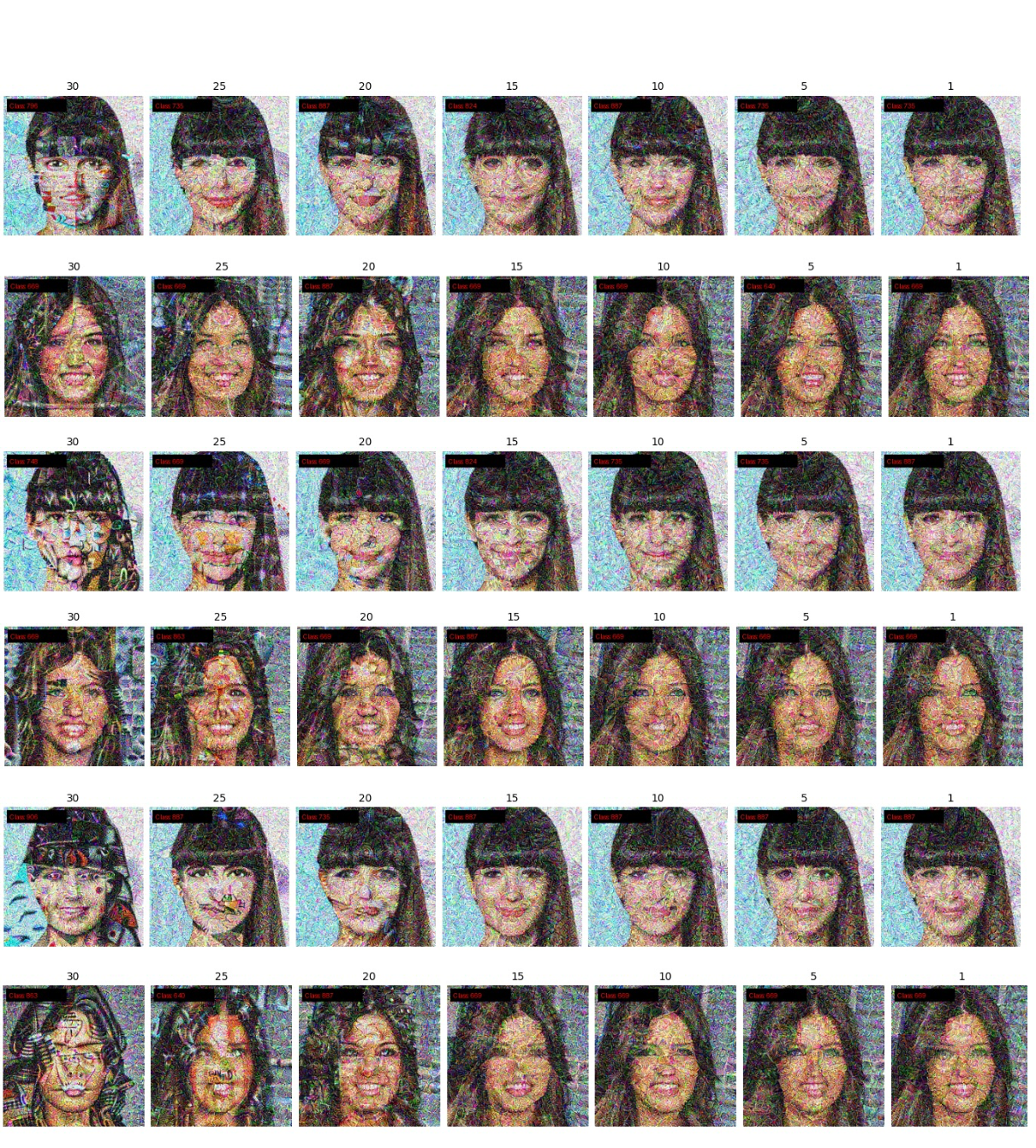}
    \caption{ Demonstrating the influence of conditional prompts in adversarial image generation based on different time steps. We implemented an adversarial attack pipeline that modifies identity classifications by leveraging both FGSM and diffusion-based adversarial sample generation. FGSM is then applied to generate adversarial noise gradually, subtly altering pixel values to mislead the classifier into assigning a different identity label.}
    \label{fig:framework}
\end{figure*}

\begin{figure}[t!] 
    \centering
    \includegraphics[width=\textwidth, height=0.9\textheight, keepaspectratio]{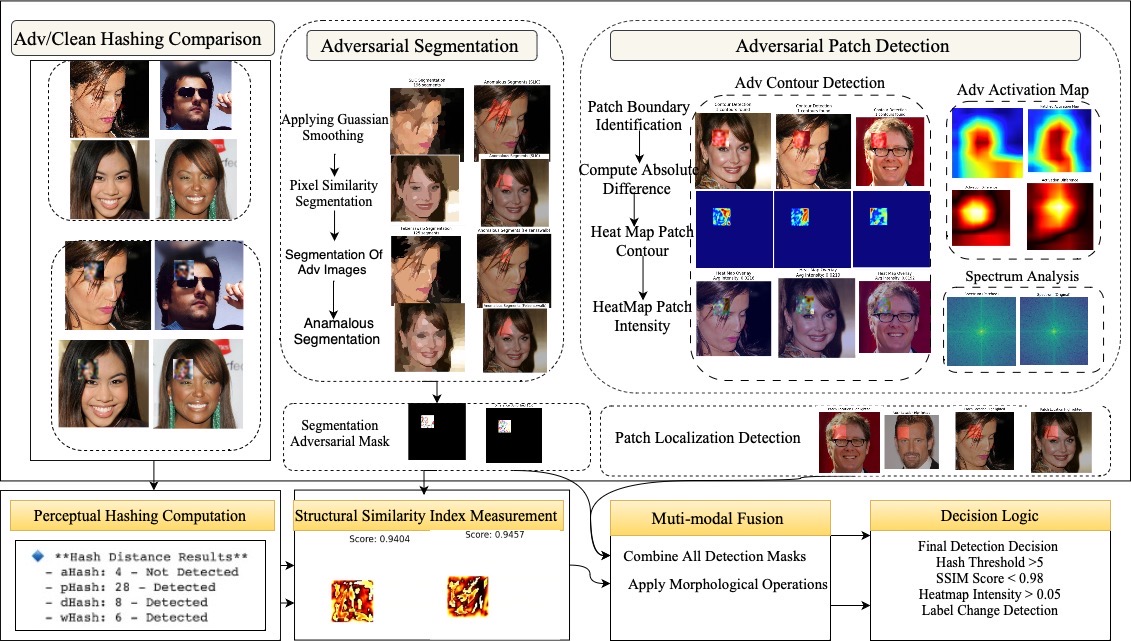}
    \caption{Overview of the proposed adversarial patch detection and localization framework. The pipeline integrates perceptual hashing, adversarial segmentation, contour- and activation-based heatmap analysis, spectral cues, and structural similarity measurements. Multi-modal fusion and threshold-based decision logic enable robust detection and precise localization of adversarial patches in facial images.}
    \label{fig:diagram0}
\end{figure}
\subsection{Perceptual Hash and Structural Similarity in Forensic Setting}
In this work, we introduce an approach for detecting adversarial patches using perceptual hashing and hash distance computation, combined with advanced image segmentation and machine learning techniques. The process begins with the preprocessing of input images, where we utilize a pre-trained ResNet50 model for image classification and a Stable Diffusion Img2Img pipeline to generate adversarial patches. The input images are resized and transformed into tensor format using standard image transformation techniques, ensuring compatibility with the neural network models in Fig.6.
\begin{figure}[h!]
    \centering
    \includegraphics[width=0.9\textwidth, height=0.4\textheight]{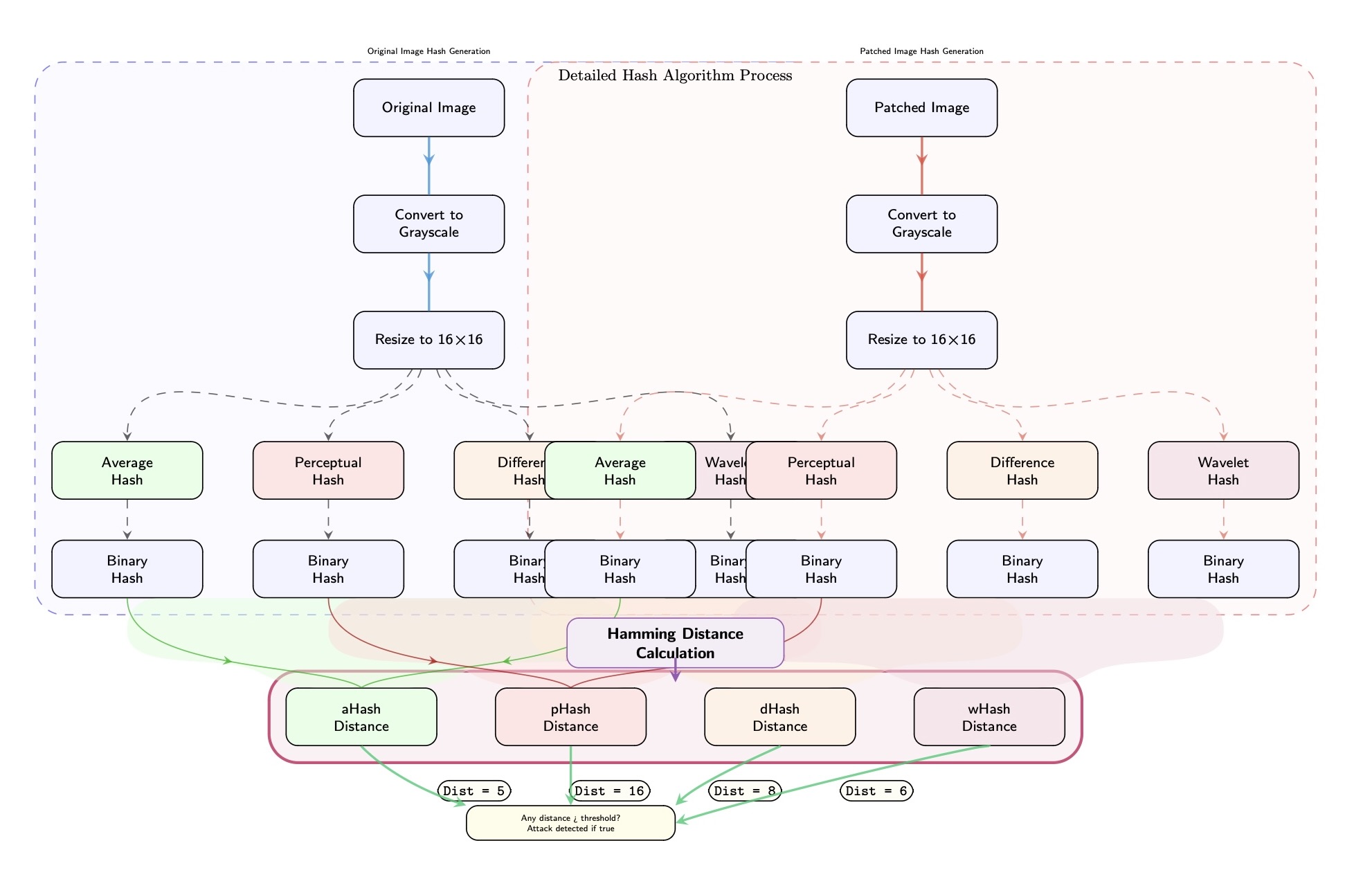}
    \caption{Visualizing a detailed hash pipeline process and hamming distance calculation.}
    \label{fig:surprise_expression}
\end{figure}

The first step in our detection methodology involves computing multiple perceptual hashes for both the original and patched images. We leverage four distinct types of hashes: \textit{aHash}, \textit{pHash}, \textit{dHash}, and \textit{wHash}. These hashing methods are designed to capture subtle differences in image structure, texture, and pixel patterns. The images are converted to grayscale, and each hash is computed using the respective method. This provides a set of hash values that serve as the foundation for detecting any potential adversarial manipulations.

For an image \( I \), the \textit{aHash} is computed by dividing the image into blocks, calculating the average color value for each block, and then producing a binary string representation where each bit represents whether the corresponding block is above or below the average.
where \( I_i \) is the color value of pixel \( i \), \( N \) is the number of pixels in the block, and \( \text{mean}(I) \) is the mean color of the image.

Similarly, the \textit{pHash} captures the perceptual content of the image by performing a discrete cosine transform (DCT) and quantizing the resulting coefficients.

The \textit{dHash} computes the differences between adjacent pixel values, while the \textit{wHash} applies wavelet transforms to capture high-frequency image components.

\subsection{Hamming Distance Calculation}

To quantify the difference between the original and patched images, we compute the \textit{Hamming distance} between the perceptual hashes. The hamming distance is defined as the number of differing bits between two binary strings in Fig.8.8. Given two perceptual hashes \( H_1 \) and \( H_2 \), the hamming distance \( d_H \) is computed as:
\[
d_H(H_1, H_2) = \sum_{i=1}^{N} \mathbb{1}\left(H_{1,i} \neq H_{2,i}\right)
\]
where \( H_{1,i} \) and \( H_{2,i} \) represent the \( i \)-th bit of hashes \( H_1 \) and \( H_2 \), respectively, and \( \mathbb{1}(\cdot) \) is the indicator function which is 1 if the condition holds and 0 otherwise.

This measure helps in identifying how much the adversarial patch has altered the image in perceptual terms. A larger hamming distance implies a greater perceptual change, potentially indicating an adversarial attack.

The next stage of analysis involves computing the \textit{SSIM} between the original and patched images. SSIM is a method for measuring the perceived quality of images by comparing luminance, contrast, and structure. It is mathematically defined as:
\[
\text{SSIM}(x, y) = \frac{(2\mu_x\mu_y + C_1)(2\sigma_{xy} + C_2)}{(\mu_x^2 + \mu_y^2 + C_1)(\sigma_x^2 + \sigma_y^2 + C_2)}
\]
where:
- \( x \) and \( y \) are the two image patches being compared.
- \( \mu_x \), \( \mu_y \) are the average pixel intensities of \( x \) and \( y \).
- \( \sigma_x^2 \), \( \sigma_y^2 \) are the variances of the pixel intensities of \( x \) and \( y \).
- \( \sigma_{xy} \) is the covariance of \( x \) and \( y \).
- \( C_1 \) and \( C_2 \) are constants to stabilize the division with weak denominators.

The SSIM score ranges from -1 (completely dissimilar) to 1 (identical), with values closer to 0 indicating more substantial differences, often due to adversarial manipulations.

To enhance the detection of localized differences, we segment both the original and patched images using advanced segmentation algorithms. Two segmentation methods are employed: \textit{Felzenszwalb segmentation} and \textit{SLIC (simple linear iterative clustering)} segmentation. These algorithms partition the image into smaller regions or segments, allowing for the analysis of specific areas that may contain the adversarial patch. By dividing the image into regions with distinct boundaries, segmentation provides a more granular analysis of the image’s structure, making it easier to identify anomalies caused by adversarial modifications.

The Felzenszwalb segmentation algorithm applies a graph-based approach, where the image is represented as a graph and a minimal spanning tree is used to identify segments based on local color similarity. Mathematically, it segments the image by minimizing an energy function that takes into account local intensity, smoothness, and the number of segments.

For the \textit{SLIC segmentation}, the image is over-segmented into superpixels based on a distance measure that considers both spatial proximity and color similarity:
\[
\text{Distance} = \sqrt{(x_1 - x_2)^2 + (y_1 - y_2)^2 + \left(\frac{C_1 - C_2}{\sigma_C}\right)^2}
\]
where \( (x_1, y_1) \) and \( (x_2, y_2) \) are the spatial coordinates of two neighboring pixels, and \( C_1 \) and \( C_2 \) are their respective color values.

Once the images are segmented, we compute the \textit{average difference} in pixel values within each segment. The segment scores are analyzed to detect anomalies—segments with significantly higher differences compared to the rest of the image are flagged as potentially containing the adversarial patch. This anomaly detection approach leverages statistical thresholds, such as \textit{two standard deviations} above the mean segment difference, to identify suspicious regions. The anomaly score for each segment \( S_k \) is given by:
\[
\text{score}(S_k) = \frac{\sum_{i \in S_k} |I_i - I'_i|}{|S_k|}
\]
where \( I_i \) and \( I'_i \) are the pixel intensities in the original and patched images, and \( |S_k| \) is the number of pixels in segment \( S_k \). This score is compared to a threshold \( \theta \), typically defined as the mean score plus two standard deviations.

In addition to segmentation, we utilize \textit{contour detection} to further localize the adversarial patch. By computing the absolute difference between the original and patched images, we identify regions where large changes have occurred. These changes are visualized as contours, which are drawn on the patched image to highlight the exact location of the adversarial patch.

To complement the contour-based detection, we generate a \textit{heatmap} to visualize the intensity of changes across the image. The heatmap is created by comparing the images in the LAB color space, which is more sensitive to color differences than the RGB space. The difference in the L, A, and B channels is combined to produce a heatmap that highlights regions of the image most affected by the adversarial patch. Gaussian blur is applied to reduce noise, and the heatmap is normalized to emphasize areas with the greatest disparity:
\[
\text{heatmap}(I, I') = \frac{|I - I'|}{\max(|I - I'|)}
\]
where \( I \) and \( I' \) are the original and patched images, respectively.
We also incorporate a \textit{neural activation map} generated by passing the image through the ResNet50 model. This map highlights the regions of the image that the model focuses on when making a classification decision. By comparing the activation maps of the original and patched images, we can assess whether the model’s attention is altered due to the adversarial patch. Differences in the activation maps can provide strong evidence of adversarial manipulation.

The activation map is derived by extracting the output of the last convolutional layer \( A(x) \) after passing the image \( x \) through the network:
\[
A(x) = f_{layer4}(x)
\]
where \( f_{layer4} \) represents the forward pass through the last convolutional block.
To visualize the results of the perceptual hash comparison, we create a \textit{hash distance bar plot}. The plot shows the hamming distance values for each of the four hash types, with a threshold line indicating the critical level at which a patch is considered detected. Bars are color-coded to differentiate between detected and undetected patches, making it easy to assess the effectiveness of each hash method in identifying adversarial changes.
Finally, the system integrates all detection methods—hash distance, SSIM, segmentation anomalies, contour detection, heatmap visualization, and neural network activation mapping—into a unified decision framework. If any of the detection methods surpass the predefined thresholds, the system flags the image as containing an adversarial patch. The results are then displayed visually, with the location of the adversarial patch highlighted, and a detailed summary of each detection method’s findings.

Through this end to end pipeline, we are able to effectively detect adversarial patches and understand their impact on image integrity. The combination of perceptual hashing, structural similarity analysis, segmentation, and neural network activation mapping provides a framework for identifying and visualizing adversarial manipulations in images in Fig.6. and Fig.7.7.
\begin{figure}[H]
    \centering
    \includegraphics[width=\textwidth, height=9\textheight, keepaspectratio]{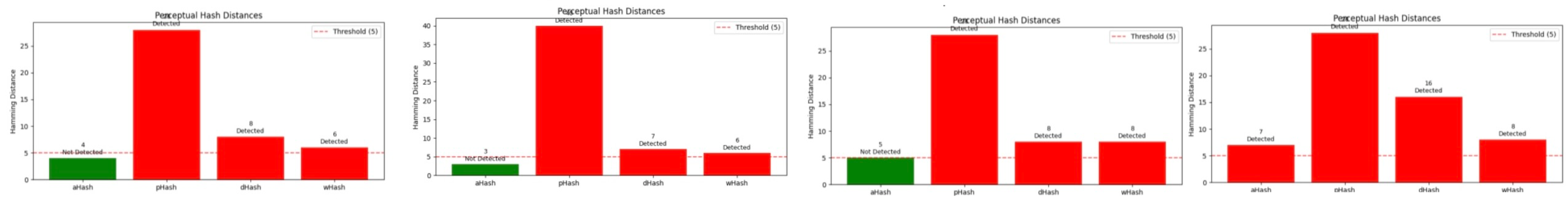}
    \caption{Displaying perceptual hash distances for four different hash methods (aHash, pHash, dHash, and wHash), with detection results based on a threshold of 5. The bars represent the hamming distances between the hashes, where the red bars indicate detection and the green bars indicate non-detection. The threshold line at 5 helps to determine which hashes exceed the detection limit.}
    \label{fig:diagram0}
\end{figure}

\section{Model Evaluation and Comparison}\label{sec12}
In this section we aim to evaluate our proposed model into different category of 1) evaluating the impact of adversarial attack for evasion attack 2) patch generation evaluation 3) impact of different parameters on attack success 4) transferable optimal adversarial patch evaluation 5) Identity evaluation for poisoned caption Identity recognition and number of detected adversarial attack detection method. 
\subsection{Evaluating the Impact of Adversarial Patches for Identity Evasion Attack}

The visualizations in Fig.9. provide compelling evidence of the effectiveness of identity patch evasion attacks on computer vision systems. The results demonstrate how adversarial patches —colorful, abstract overlays on facial images successfully disrupt facial recognition and depth perception mechanisms.
These carefully crafted patches in facial recognition conceal identities while preserving human recognizability, creating a disconnect between machine and human perception. The identity patch evasion Attack results illustrate this effect, where the patched faces evade recognition while remaining visibly identifiable to humans. This highlights the fundamental vulnerability of AI-based recognition systems when confronted with adversarial manipulations.
Even more concerning is the impact on depth perception. The depth monocular sensing visualization section reveals drastic distortions in the perceived spatial structure of faces. Original depth maps depict smooth facial surfaces, while patched depth maps introduce abrupt geometric anomalies. The difference maps, marked by intense red and yellow regions, further expose the severity of these distortions, suggesting interference with the model’s depth estimation process.
The 3D depth map comparison before and after patch evasion attack results in Fig.9. further quantify these disruptions. The original depth maps exhibit well-defined facial structures, while the patched depth maps introduce unnatural protrusions and depressions. The depth difference maps, with their pronounced spikes and valleys, quantify these estimation errors, demonstrating how the attack method persistently destabilizes spatial understanding across different samples.
\begin{figure*}[t!]
    \centering
    \includegraphics[width=0.9\textwidth, height=0.4\textheight, keepaspectratio]{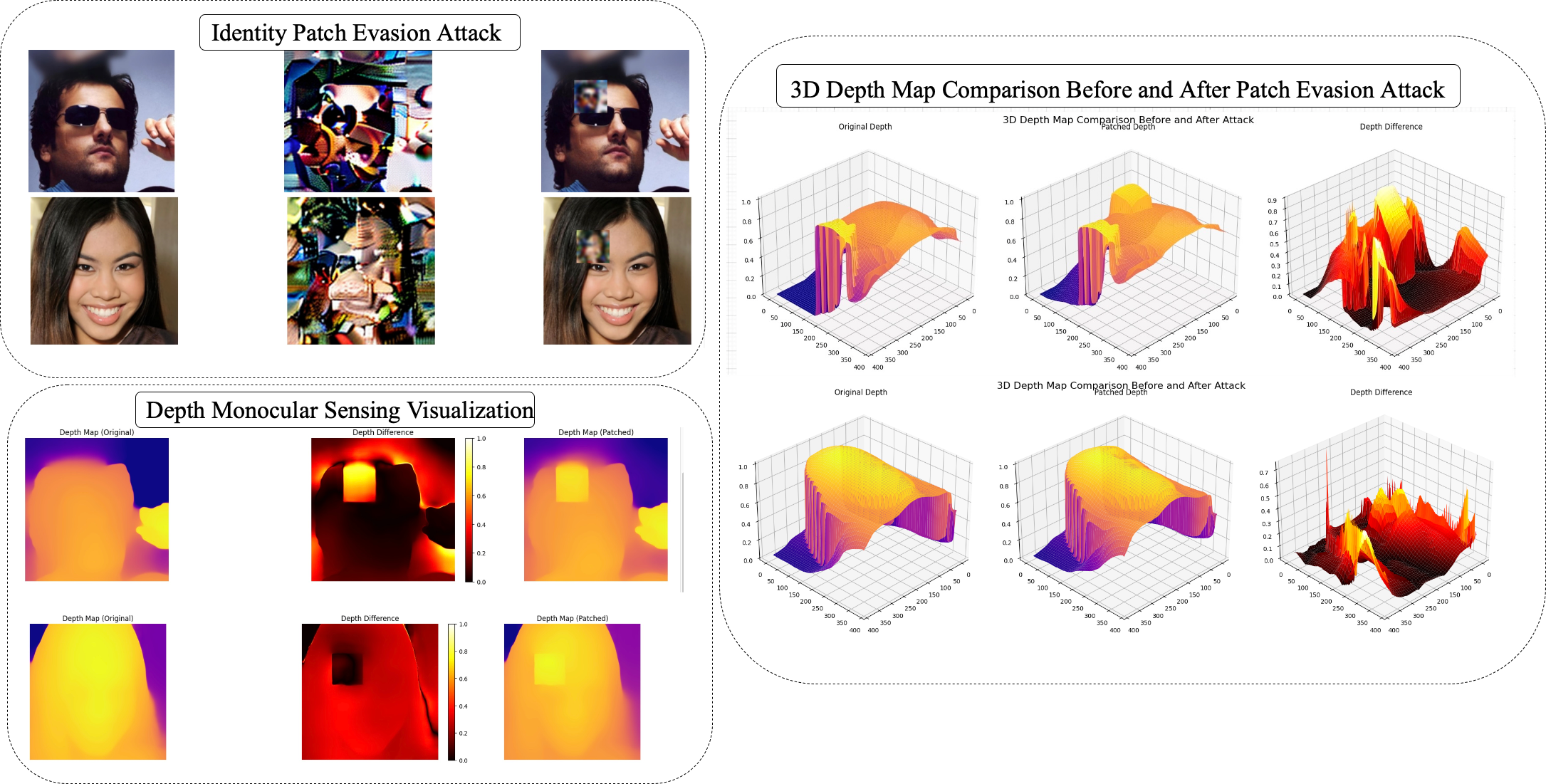}
    \caption{Effects of identity patch evasion attacks on facial recognition and depth perception.}
    \label{fig:framework}
\end{figure*}
\subsection{Patch Generation Evaluation}
The table titled \textit{Comparison of proposed work with other studies} compares the performance of adversarial patches across multiple studies, using key metrics such as SSIM, LPIPS, L2 distance, MS-SSIM, transferability, and overall score.

Our proposed work pre-trained models for identity classification, with LPIPS used to assess perceptual quality. The diffusion model is used to refine adversarial patches through diffusion-based denoising, improving their imperceptibility. Adversarial patches are generated using the FGSM and then refined for robustness.

Several metrics are used for evaluation: SSIM measures structural similarity,LPIPS assesses perceptual quality,MS-SSIM evaluates multiple image scales, and L2 distance quantifies pixel-wise differences. Transferability is tested by evaluating patch success across models.
The results are tested with various patch configurations (diffusion strength, size, and position). Study 1 \cite{zhou2019scene} shows similar performance but with slightly lower SSIM and MS-SSIM scores, and higher transferability. Study 2\cite{xie2017resnext} achieves comparable LPIPS and SSIM scores but has slightly lower MS-SSIM. Study 3 \cite{szegedy2014intriguing} shows lower scores across all metrics, indicating less effectiveness in fooling models.

Overall, the proposed work demonstrates strong performance, with a balanced approach that ensures high imperceptibility, effectiveness, and transferability, making it a competitive method in adversarial patch generation in Table 1.

\begin{table}[ht]
\caption{Comparison of Metrics and Performance Across Different Studies. The changes are indicated with arrows showing the direction of increase or decrease.}
\label{tab:comparison_metrics_performance}
\centering
\begin{tabular*}{\textwidth}{@{\extracolsep\fill}lcccccc}
\toprule
& \multicolumn{3}{@{}c@{}}{Metrics} & \multicolumn{3}{@{}c@{}}{Performance} \\
\cmidrule{2-4} \cmidrule{5-7}
\textbf{Sample} & \textbf{SSIM} & \textbf{LPIPS} & \textbf{L2} & \textbf{MS-SSIM} & \textbf{Transfer} & \textbf{Score} \\
\midrule
Proposed Work & 0.94 \textcolor{green}{\(\uparrow\)} & 0.090 \textcolor{green}{\(\downarrow\)} & 0.078 \textcolor{green}{\(\downarrow\)} & 0.808 \textcolor{green}{\(\uparrow\)} & 0.389 \textcolor{green}{\(\downarrow\)} & 0.705 \textcolor{green}{\(\uparrow\)} \\
Study 1       & 0.92 \textcolor{red}{\(\downarrow\)} & 0.080 \textcolor{green}{\(\downarrow\)} & 0.070 \textcolor{green}{\(\downarrow\)} & 0.790 \textcolor{green}{\(\downarrow\)} & 0.400 \textcolor{green}{\(\uparrow\)} & 0.685 \textcolor{red}{\(\downarrow\)} \\
Study 2       & 0.93 \textcolor{green}{\(\uparrow\)} & 0.085 \textcolor{red}{\(\uparrow\)} & 0.065 \textcolor{green}{\(\downarrow\)} & 0.805 \textcolor{green}{\(\uparrow\)} & 0.420 \textcolor{green}{\(\uparrow\)} & 0.700 \textcolor{green}{\(\uparrow\)} \\
Study 3       & 0.91 \textcolor{red}{\(\downarrow\)} & 0.095 \textcolor{red}{\(\uparrow\)} & 0.085 \textcolor{red}{\(\uparrow\)} & 0.775 \textcolor{red}{\(\downarrow\)} & 0.375 \textcolor{red}{\(\downarrow\)} & 0.670 \textcolor{red}{\(\downarrow\)} \\
\botrule
\end{tabular*}
\end{table}

\subsection{Impact of Different Parameters on Attack Success, Confidence Drop, and Effectiveness Score}
Figure 10 consists of three bar charts that provide a detailed comparison of different parameters influencing the performance of an attack. The first chart, titled , demonstrates the success rate of the attack across six different parameters: target\_class, epsilon, position, size, blur\_radius, and diffusion\_strength. It is evident from the chart that parameters like target\_class, epsilon, and diffusion\_strength exhibit a high success rate of around 80\%, while position and size contribute to a noticeably lower success rate of 60\%. 

The second chart shows the average confidence drop across the same parameters. Diffusion\_strength again stands out with the highest drop in confidence, followed by target\_class, which also demonstrates a relative drop. The other parameters, including epsilon, position, and size, have a smaller impact on the confidence drop, reflecting a consistent trend with the attack success rate chart where fewer changes were observed in these parameters.

Finally, the third chart, provides an assessment of the best combined score of each parameter, illustrating how each one contributes to the overall attack effectiveness. Target\_class stands out with the highest score of 2.86, while position and size perform relatively poorly with scores around 2.78. This reinforces the conclusion that some parameters, such as target\_class and diffusion\_strength, lead to a more effective attack compared to others.

\begin{figure}[H]
    \centering
    \includegraphics[width=0.8\textwidth, height=0.7\textheight, keepaspectratio]{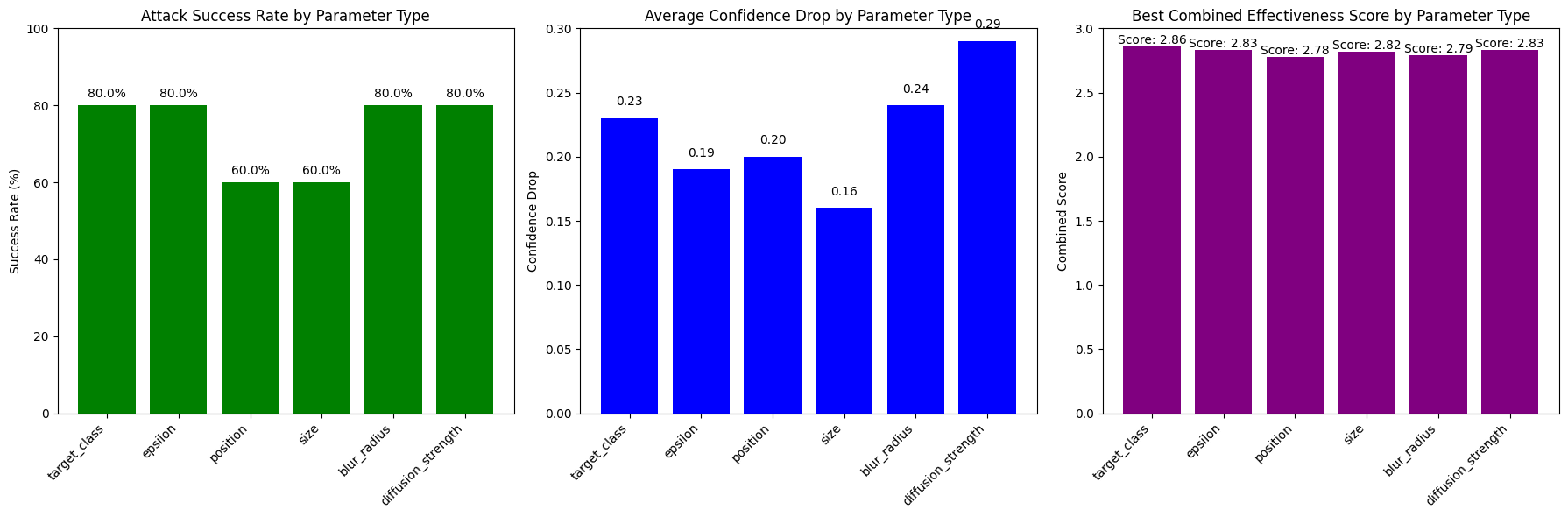}
    \caption{Analysis of adversarial patch attack performance: success rate, confidence drop, and combined effectiveness score across parameters like target class, epsilon, position, size, blur radius, and diffusion strength.}
    \label{fig:diagram0}
\end{figure}

\subsection{Transferable Optimal Adversarial Patch Evaluation}
Figure 11 focuses on patch variation analysis and examines how position variation affects our adversarial attack success. Our left side shows the original image, while subsequent images display different patch positions with their class and confidence scores. Our confidence drop metrics demonstrate how the attack's effectiveness changes as the patch moves across the image. Some positions in our analysis yield a higher success rate (with confidence drops of 0.03 and 0.10), while others diminish our attack's success (with larger drops of up to 0.21) demonstrating how patch positioning influences effectiveness. Our second figure presents an optimal adversarial patch analysis featuring our best-performing patch. It displays our original image alongside our optimal adversarial patch. Our patched image demonstrates the attack's success through a marked shift in image classification, evidenced by changes in confidence scores. This shows how effectively our patch manipulates the model's classification. Our difference visualization highlights the subtle changes between our original and patched images that alter the model's prediction.  In Table 2, we evaluated our proposed model for testing the transferability attribute of adversarial patch generation and their respective quality for further examination for different groups of samples based on different geographical places of adversarial patch creation of Fig.12.

\begin{figure}[H]
    \centering
    \includegraphics[width=0.7\textwidth, height=0.2\textheight, keepaspectratio]{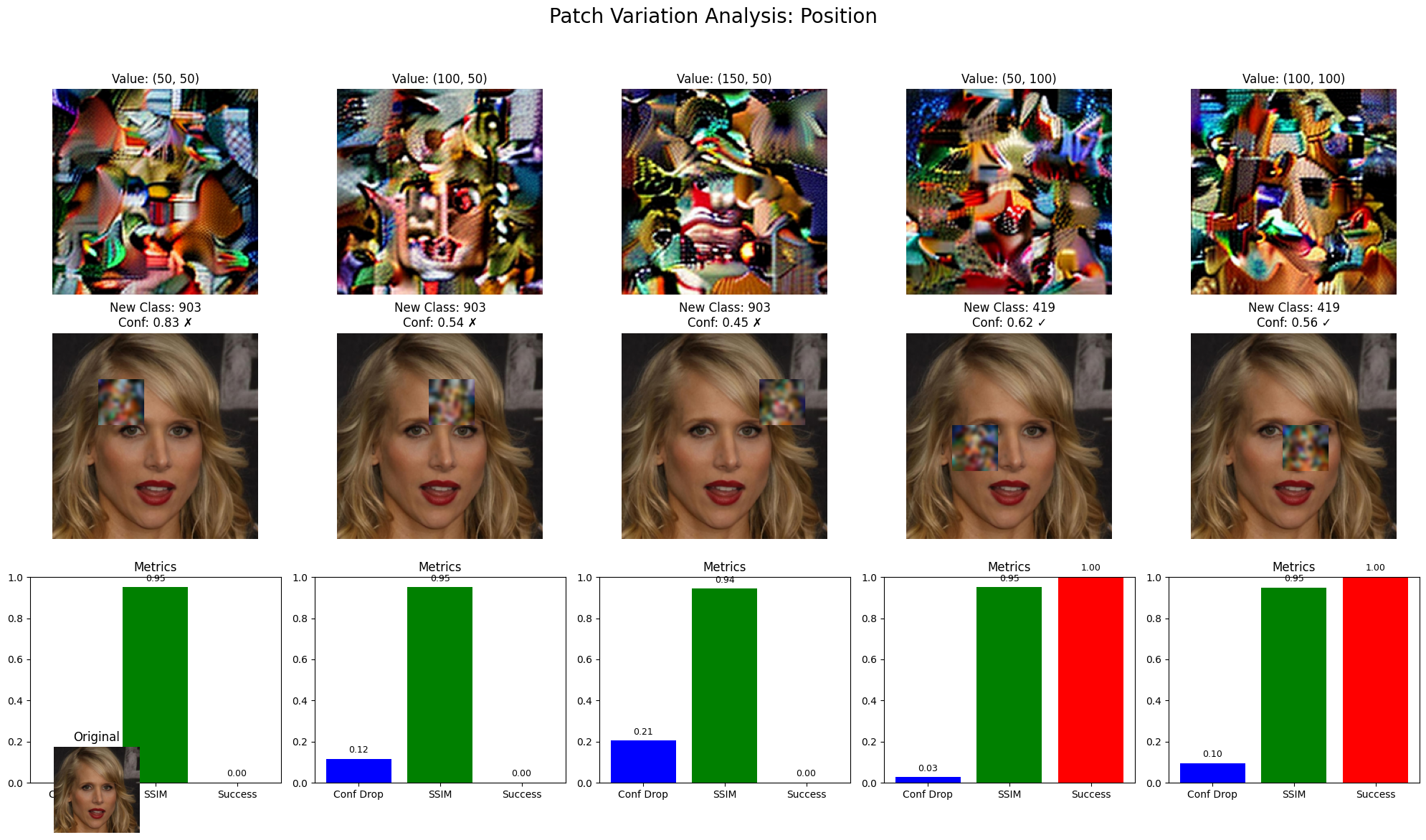}
    \caption{Patch variation analysis for different positions. The images show the impact of adversarial patch positions on classification results, with metrics displaying SSIM, confidence drop, and success rates.}
    \label{fig:diagram0}
\end{figure}
\vspace{-40pt}
\begin{figure}[H]
    \centering
    \includegraphics[width=0.7\textwidth, height=0.3\textheight, keepaspectratio]{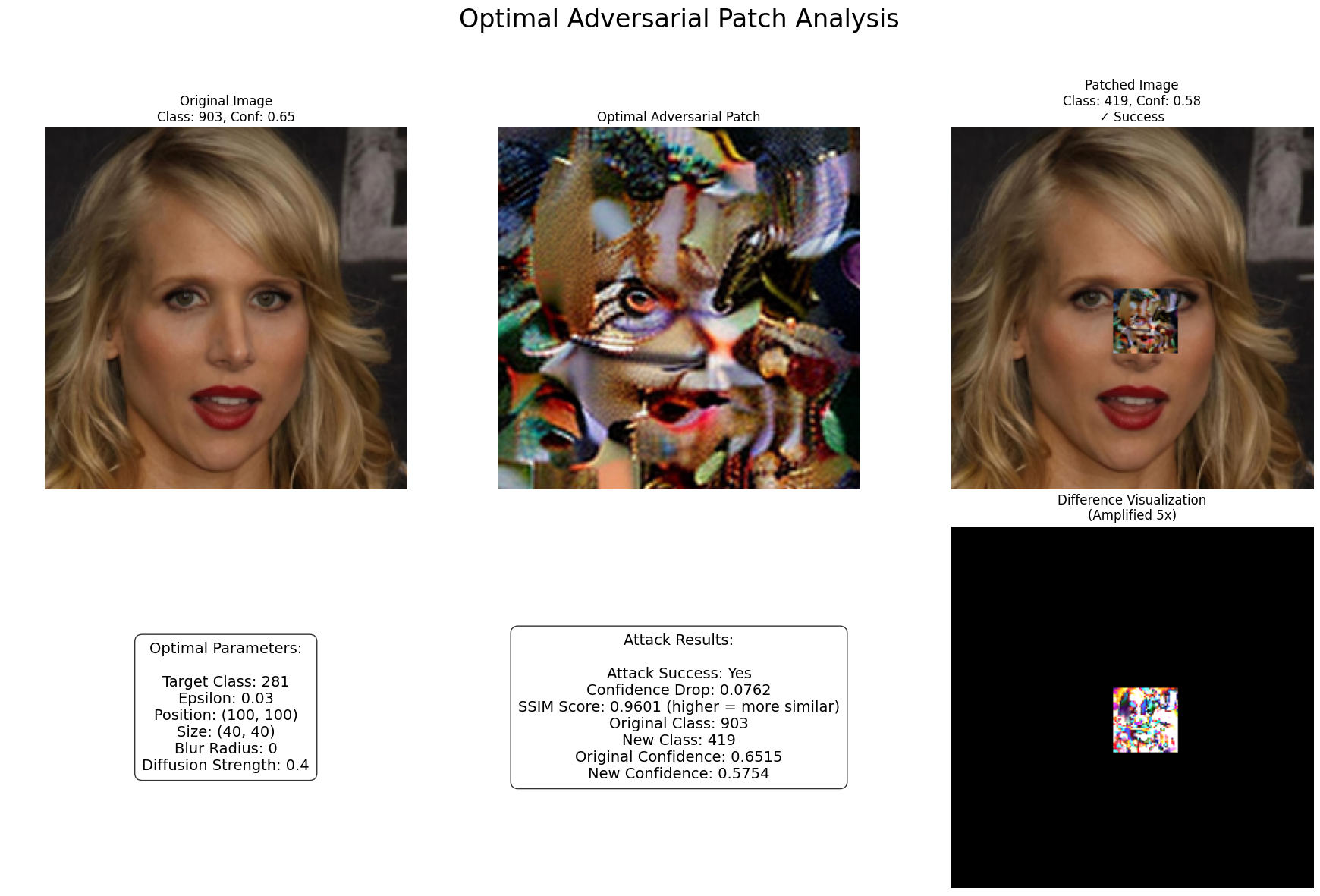}
    \caption{The images show the original (left), optimal adversarial patch (center), and patched image (right), demonstrating a successful attack with changes in classification, confidence drop, SSIM score, and attack parameters.}
    \label{fig:diagram1}
\end{figure}
\begin{table}[h]
\caption{Metrics for Adversarial Patch Transferability Testing across Various Configurations and Evaluation. Success is indicated by "Yes" and "No", and transferability varies from 0 to 1 across samples.}\label{tab:attack_results}
\begin{tabular*}{\textwidth}{@{\extracolsep\fill}lcccccc}
\toprule
& \multicolumn{3}{@{}c@{}}{Metrics for Position (50, 50)} & \multicolumn{3}{@{}c@{}}{Metrics for Position (100, 100)} \\
\cmidrule{2-4} \cmidrule{5-7}
Sample & Success & Confidence drop & SSIM & Success & Confidence drop & SSIM \\
\midrule
Group 1 & Yes & -0.1507 & 0.9777 & Yes & -0.0855 & 0.9812 \\
Group 2 & No & 0.0529 & 0.9798 & Yes & -0.0824 & 0.9771 \\
Group 3 & Yes & 0.0974 & 0.9827 & Yes & 0.0627 & 0.9805 \\
Group 4 & Yes & -0.0541 & 0.9759 & Yes & -0.5963 & 0.9765 \\
Group 5 & Yes & 0.1017 & 0.9801 & Yes & -0.0336 & 0.9754 \\
\midrule
& \multicolumn{3}{@{}c@{}}{LPIPS Metrics} & \multicolumn{3}{@{}c@{}}{Transferability} \\
\cmidrule{2-4} \cmidrule{5-7}
Sample & LPIPS & Transferability & LPIPS & Transferability & LPIPS & Transferability \\
\midrule
Group 1 & 0.0444 & 0.50 & 0.0256 & 0.50 & 0.0297 & 0.75 \\
Group 2 & 0.0176 & 0.50 & 0.0261 & 1.00 & 0.0264 & 0.75 \\
Group 3 & 0.0326 & 0.25 & 0.0256 & 0.50 & 0.0235 & 0.25 \\
Group 4 & 0.0370 & 0.25 & 0.0336 & 0.75 & 0.0581 & 0.50 \\
Group 5 & 0.0295 & 0.25 & 0.0505 & 0.50 & 0.0385 & 0.00 \\
\midrule
\multicolumn{2}{@{}l@{}}{\textbf{Overall Attack Success Rate:}} & \multicolumn{2}{@{}c@{}}{81.11\%} \\
\multicolumn{2}{@{}l@{}}{\textbf{Average Metrics:}} & \multicolumn{2}{@{}c@{}}{Confidence Drop: -0.1696} \\
\multicolumn{2}{@{}l@{}}{} & \multicolumn{2}{@{}c@{}}{SSIM Score: 0.9403} \\
\multicolumn{2}{@{}l@{}}{} & \multicolumn{2}{@{}c@{}}{LPIPS Score: 0.0840} \\
\multicolumn{2}{@{}l@{}}{} & \multicolumn{2}{@{}c@{}}{MS-SSIM Score: 0.8664} \\
\multicolumn{2}{@{}l@{}}{} & \multicolumn{2}{@{}c@{}}{Transferability: 0.6222} \\
\midrule
\multicolumn{2}{@{}l@{}}{\textbf{Best Configuration:}} & \multicolumn{2}{@{}c@{}}{Sample: 2} \\
\multicolumn{2}{@{}l@{}}{} & \multicolumn{2}{@{}c@{}}{Diffusion Strength: 0.5} \\
\multicolumn{2}{@{}l@{}}{} & \multicolumn{2}{@{}c@{}}{Patch Size: 30} \\
\multicolumn{2}{@{}l@{}}{} & \multicolumn{2}{@{}c@{}}{Position: (150, 50)} \\
\multicolumn{2}{@{}l@{}}{} & \multicolumn{2}{@{}c@{}}{Success: Yes} \\
\multicolumn{2}{@{}l@{}}{} & \multicolumn{2}{@{}c@{}}{Composite Score: 0.9927} \\
\botrule
\end{tabular*}
\end{table}

\subsection{Identity Evaluation}
In our comprehensive approach for generating adversarial captions, we evaluate their impact through various strategies, including semantic and grammatical alterations, contextual distortions, and other transformations. This evaluation delves deeply into the robustness of automated captioning systems. We leverage pre-trained models such as vision encoder decoder model and ViT image processor to generate captions from images. Following this, we apply a series of adversarial strategies to modify these captions, allowing us to assess the impact these modifications have on the integrity and effectiveness of the generated content. Our approach employs a wide range of metrics, including BLEU, METEOR, ROUGE, and accuracy metrics, to evaluate both clean and adversarial captions.

In our implementation, we developed the adversarial caption generator class, which plays a central role by applying a series of poisoning strategies to the generated captions. We designed these strategies to involve significant alterations that make the captions adversarial. For instance, we use semantic distortion to replace words with synonyms, grammatical mutation to modify the grammatical structure, and contextual hallucination to introduce unexpected elements into the caption.

We have made the use of poisoning templates a crucial feature of our adversarial caption generator. We designed these templates to flag potential poisoned captions by applying structured, contextually relevant modifications to the original captions. Rather than introducing random changes, we use the templates to guide the modification of specific elements such as nouns, adjectives, and verbs.

In our implementation, we developed the caption dataset class to support efficient processing of images and captions. We designed it to facilitate the loading of images, generation of captions through pre-trained models like ViT-GPT2, and application of adversarial transformations.

In our evaluation of the adversarial captions, we found several important results. We observed BLEU-1 and BLEU-4 scores of 0.5377 and 0.2741, respectively, showing us a moderate overlap in n-grams between the original and adversarial captions. Our METEOR score of 0.6473 reflects a reasonable alignment between our generated and human-written references. We also found that our ROUGE scores (ROUGE-1 at 0.6967, ROUGE-2 at 0.3131, and ROUGE-L at 0.4666) further support these findings. Through our clean accuracy score of 0.9231, we confirmed that a significant portion of the original captions remained intact, while our adversarial accuracy score of 0.4069 showed us that the adversarial captions achieved moderate deviation from the original content in Table 3.
\begin{table}[h]
\caption{Caption Metrics Results}
\label{tab1}%
\begin{tabular}{@{} p{4cm} p{4cm} @{}}
\toprule
\textbf{Metric} & \textbf{Value} \\ 
\midrule
BLEU-1          & 0.5377        \\ 
BLEU-4          & 0.2741        \\ 
METEOR          & 0.6473        \\ 
ROUGE-1         & 0.6967        \\ 
ROUGE-2         & 0.3131        \\ 
ROUGE-L         & 0.4666        \\ 
Clean accuracy  & 0.9231        \\ 
Adversarial accuracy & 0.4069    \\ 
Semantic shift  & 0.4157        \\ 
\bottomrule
\end{tabular}
\footnotetext{Source: Results obtained from adversarial caption generation evaluation metrics.}
\end{table}

\subsection{Adversarial Detection}
Figure 13 presents analysis of SSIM scores, which are used to assess the visual similarity between two images. The first plot on the left shows the distribution of SSIM scores for 500 images, with a clear threshold (0.98) marked by a red dashed line. This threshold is set to help determine the detection rate, which in this case is 100\%, indicating that all images have been successfully detected based on this threshold.

The second plot in the center displays the sorted SSIM scores for the 500 images, ordered from lowest to highest. The threshold is again marked with a red dashed line at 0.98, providing a reference point for the sorted scores. This plot visually confirms that all 500 images have SSIM scores higher than the threshold, with a detection rate of 100\%.

The third plot, on the right, presents a box plot of SSIM scores, offering a summary of the statistical distribution. The plot shows the minimum, first quartile (Q1), median, third quartile (Q3), and maximum SSIM values, providing insights into the range and variability of the scores. The SSIM scores are concentrated around the 0.946 median, with a narrow interquartile range, indicating high similarity between the images.

Together, these visualizations provide a detailed overview of the SSIM score distribution and detection performance, demonstrating the consistency and effectiveness of the thresholding method for image comparison Table 4.
\begin{figure}[H]
    \centering
    \includegraphics[width=\textwidth, height=9\textheight, keepaspectratio]{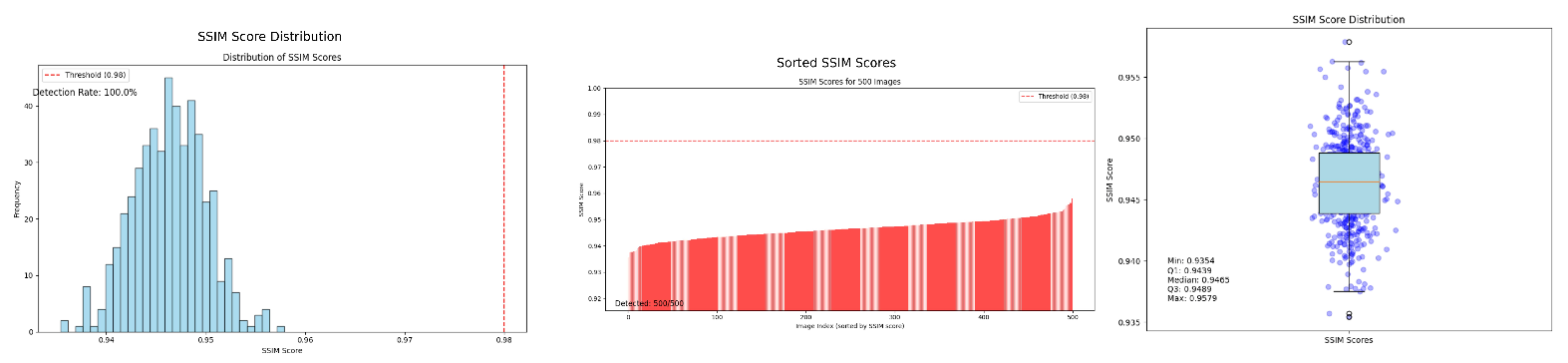}
    \caption{Demonstrating the number of detected adversaries from an imbalance dataset by visualization and tracking the SSIM distribution}
    \label{fig:diagram0}
\end{figure}
\begin{table}[ht]
\centering
\caption{SSIM Score Distribution Summary}
\label{tab:ssim_summary}
\begin{tabular}{|c|c|}
\hline
\textbf{Metric} & \textbf{Value} \\ \hline
\textbf{Minimum SSIM score} & 0.9354 \\ \hline
\textbf{Q1 (First quartile)} & 0.9419 \\ \hline
\textbf{Median SSIM Score} & 0.9465 \\ \hline
\textbf{Q3 (Third quartile)} & 0.9489 \\ \hline
\textbf{Maximum SSIM score} & 0.9579 \\ \hline
\textbf{Threshold for detection} & 0.98 \\ \hline
\textbf{Detection rate} & 100\% \\ \hline
\end{tabular}
\end{table}

\section{Conclusion}\label{sec13}
In this scope of study, we thoroughly examine this forensic occurrence of adversaries. We witnessed and explored the opportunities of adversarial patch generation using a stable diffusion model, and we extended our methodology to a systematic approach design to detect and poison the number of identity recognition of adversarial patch attacks to the dataset. We have successfully detected and analyzed the respected measurement through this study. We achieved excellent evaluation metrics by evaluating our proposed methodology and detecting adversarial patch attacks. 

\backmatter


\bmhead{Acknowledgments}
The research was partially sponsored by the Army Research Office and was carried out under Grant Number W911NF-21-1-0264. The views and conclusions contained in this document are those of the authors and should not be interpreted as representing the official policies, either expressed or implied, of the Army Research Office or the US Government. The US government is authorized to reproduce and distribute reprints for Government purposes, notwithstanding any copyright notation herein. \\
In addition, this research was supported as part of the 2024 Summer Research Experience for Graduate Students Program of the Northrop Grumman Research and Education Program (NG REP) through the Florida A\& M University Foundation
This arXiv submission is a preprint. A revised version of this work has been accepted for publication in the Springer Nature book AI-Driven Forensics. This version includes one additional figure for completeness. 

\bibliography{sn-bibliography}




\end{document}